\documentclass[11pt,letterpaper]{article}
\usepackage{emnlp2016}
\usepackage{times}
\usepackage{latexsym}

\emnlpfinalcopy

\def\emnlppaperid{345}

\title{Improving Multilingual Named Entity Recognition \\with Wikipedia Entity Type Mapping}

\author{Jian Ni \and Radu Florian\\
    IBM T. J. Watson Research Center\\
    1101 Kitchawan Road, Yorktown Heights, NY 10598, USA\\
  {\tt \{nij, raduf\}@us.ibm.com}}

\date{}

\begin{document}

\maketitle

\begin{abstract}
The state-of-the-art named entity recognition (NER) systems are statistical machine learning models that have strong generalization capability (i.e., can recognize unseen entities that do not appear in training data) based on lexical and contextual information. However, such a model could still make mistakes if its features favor a wrong entity type. In this paper, we utilize Wikipedia as an open knowledge base to improve multilingual NER systems. Central to our approach is the construction of high-accuracy, high-coverage multilingual Wikipedia entity type mappings. These mappings are built from weakly annotated data and can be extended to new languages with no human annotation or language-dependent knowledge involved. Based on these mappings, we develop several approaches to improve an NER system. We evaluate the performance of the approaches via experiments on NER systems trained for 6 languages. Experimental results show that the proposed approaches are effective in improving the accuracy of such systems on unseen entities, especially when a system is applied to a new domain or it is trained with little training data (up to 18.3 $F_1$ score improvement).
\end{abstract}

\section{Introduction}

Named entity recognition (NER) is an important NLP task that automatically detects entities in text and classifies them into pre-defined entity types such as persons, organizations, geopolitical entities, locations, events, etc. NER is a fundamental component of many
information extraction and knowledge discovery applications, including relation extraction, entity linking, question answering and data mining.

The state-of-the-art NER systems are usually statistical machine learning models that are trained with human-annotated data. Popular models include maximum entropy Markov models (MEMM) \cite{mccallum00}, conditional random fields (CRF) \cite{lafferty01} and neural
networks \cite{collobert11,lample16}. Such models have strong generalization capability to recognize \emph{unseen entities}\footnote{An entity is an \emph{unseen entity} if it does not appear in the training data used to train the NER model.} based on lexical and contextual information (features). However, a model could still make mistakes if its features favor a wrong entity type, which happens more frequently for unseen entities as we have observed in our experiments.

Wikipedia is an open-access, free-content Internet encyclopedia, which has become the de facto on-line source for general reference. A Wikipedia page about an entity normally includes both structured information and unstructured text information, and such information can
be used to help determine the entity type of the referred entity.

So far there are two classes of approaches that exploit Wikipedia to improve NER. The first class of approaches use Wikipedia to generate features for NER systems, e.g., \cite{kazama07,ratinov09,radford15}. \newcite{kazama07} try to find the Wikipedia entity for each candidate word sequence and then extract a category label from the first sentence of the Wikipedia entity page. A part-of-speech (POS) tagger is used to extract the category label features in the training and decoding phase. \newcite{ratinov09} aggregate several Wikipedia categories into higher-level concept and build a gazetteer on top of it. The two approaches were shown to be able to improve an English NER system. Both approaches, however, are language-dependent because \cite{kazama07} requires a POS tagger and \cite{ratinov09} requires manual category aggregation by inspection of the annotation guidelines and the training set.  \newcite{radford15} assume that document-specific knowledge base (e.g., Wikipedia) tags for each document are provided, and they use those tags to build gazetteer type features for improving an English NER system.

The second class of approaches use Wikipedia to generate weakly annotated data for training multilingual NER systems, e.g., \cite{richman08,nothman13}. The motivation is that annotating multilingual NER data by human is both expensive and time-consuming. \newcite{richman08} utilize the category information of Wikipedia to determine the entity type of an entity based on manually constructed rules (e.g., category phrase ``Living People" is mapped to entity type PERSON). Such a rule-based entity type mapping is limited both in accuracy and coverage, e.g., \cite{toral06}. \newcite{nothman13} train a Wikipedia entity type classifier using human-annotated Wikipedia pages. Such a supervised-learning based approach has better accuracy and coverage, e.g., \cite{dakka08}. A number of heuristic rules are developed in both works to label the Wikipedia text to create weakly annotated NER training data. The NER systems trained with the weakly annotated data may achieve similar accuracy compared with systems trained with little human-annotated data (e.g., up to 40K tokens as in \cite{richman08}), but they are still significantly worse than well-trained systems (e.g., a drop of 23.9 $F_1$ score on the CoNLL data and a drop of 19.6 $F_1$ score on the BBN data as in \cite{nothman13}).

In this paper, we propose a new class of approaches that utilize Wikipedia to improve multilingual NER systems. Central to our approaches is the construction of high-accuracy, high-coverage multilingual Wikipedia entity type mappings. We use weakly annotated data
to train an English Wikipedia entity type classifier, as opposed to using human-annotated data as in \cite{dakka08,nothman13}. The accuracy of the classifier is further improved via self-training. We apply the classifier on all the English Wikipedia pages and
construct an English Wikipedia entity type mapping that includes entities with high classification confidence scores. To build multilingual Wikipedia entity type mappings, we generate weakly annotated classifier training data for another language via projection
using the inter-language links of Wikipedia. This approach requires no human annotation or language-dependent knowledge, and thus can be easily applied to new languages.

Our goal is to utilize the Wikipedia entity type mappings to improve NER systems. A natural approach is to use a mapping to create dictionary type features for training an NER system. In addition, we develop several other approaches. The first approach applies an entity type mapping as a \emph{decoding constraint} for an NER system. The second approach uses a mapping to \emph{post-process} the output of an NER system. We also design a robust \emph{joint} approach that combines the decoding constraint approach and the post-processing approach in a smart way. We evaluate the performance of the Wikipedia-based approaches on NER systems trained for 6 languages. We find that when a system is well trained (e.g., with 200K to 300K tokens of human-annotated data), the dictionary feature approach achieves the best improvement over the baseline system; while when a system is trained with little human-annotated training data (e.g., 20K to 30K tokens), a more aggressive decoding constraint approach achieves the best improvement. In both scenarios, the Wikipedia-based approaches are effective in improving the accuracy on unseen entities, especially when a system is applied to a new domain (3.6 $F_1$ score improvement on political party articles/English NER) or it is trained with little training data (18.3 $F_1$ score improvement on Japanese NER).

We organize the paper as follows. We describe how to build English Wikipedia entity type mapping in Section 2 and extend it to multilingual mappings in Section 3. We present several Wikipedia-based approaches for improving NER systems in Section 4 and evaluate their performance in Section 5. We conclude the paper in Section 6.

\section{English Wikipedia Entity Type Mapping}

In this section, we focus on English Wikipedia. We divide Wikipedia pages into two types:

\begin{itemize}

\item Entity pages that describe an entity or object, either a named entity such as ``Michael Jordan" or a common entity such as ``Basketball."

\item Non-entity pages that do not describe a certain entity, including disambiguation pages, redirection pages, list pages, etc.

\end{itemize}

We have developed an in-house English NER system \cite{florian04}. The system has 51 entity types, and the main motivation of deploying such a fine-grained entity type set is to build cognitive question answering applications on top of the NER system. An important check for a question answering system is the capability to detect whether a particular answer matches the expected type derived from the question. The entity type system used in this paper has been engineered to cover many of the frequent types that are targeted by naturally-phrased questions (such as PERSON, ORGANIZATION, GPE, TITLEWORK, FACILITY, EVENT, DATE, TIME, LOCATION, etc), and it was created over a long period of time, being updated as more types were found to be useful for question answering, and to improve inter-annotator consistency.

We want to classify Wikipedia pages into one of the entity types used in the NER system. For non-entity pages and entity pages describing common entities, we assign them with a new type OTHER.

\subsection{Wikipedia Entity Type Classification}

\subsubsection{Features}

We build maximum entropy classifiers \cite{nigam99} for Wikipedia entity type classification. We use both structured information and unstructured information of a Wikipedia page as features.

Each Wikipedia page has a unique \emph{title}. The title of an entity page is usually the name of the entity, and may include auxiliary information in a bracket to distinguish entities with the same name. We use both the entity name and auxiliary information in a bracket (if any) of a Wikipedia title as features because each could provide useful information for entity type classification. For example, based on the word ``Prize" in the title ``Nobel Prize" or the word ``Awards" in the title ``Academy Awards", one can infer that the entity type is AWARD. Likewise, the auxiliary information ``company" in the title ``Jordan (company)" indicates that the entity is an ORGANIZATION, and the auxiliary information ``film" in the title ``Alien (film)" indicates that the entity is a TITLEWORK.

The \emph{text} in a Wikipedia page of an entity provides rich information about the entity. A person can usually correctly infer the entity type by reading the first few sentences of the text in a Wikipedia page. Using more sentences provides additional information about the
entity which might be helpful, but it is also more likely to introduce noisy information which could affect the classification accuracy adversely. Therefore, we use the first 200 tokens of the text in a Wikipedia page and create $n$-gram word features out of them. We have also found that including additional $n$-gram word features of the \emph{first sentence} in a Wikipedia page results in a better classification accuracy.

Most Wikipedia pages also have a structured table called \emph{infobox}, which is placed on the right top of a page. An infobox contains attribute-value pairs, often providing summary information about an entity. The attributes in an infobox could be particularly useful for entity type classification. For example, the attribute ``Born" in an infobox provides strong evidence that the corresponding entity is a PERSON; and the attribute ``Headquarters" implies that the corresponding entity is an ORGANIZATION. We include the infobox attributes as classifier features.

\subsubsection{Training and Test Data} \label{section:eng-data}

Entity linking (EL) or entity disambiguation is the task of determining the identities of entities mentioned in text, by linking each entity to an entry (if exists) in an open knowledge base such as Wikipedia \cite{han11,hoffart11}. We apply an EL system \cite{sil14} to generate training data for Wikipedia entity type classification as follows: if a named entity in our NER training data with entity type $T$ is linked to a Wikipedia page, that page will be labeled with entity type $T$. Similarly, we apply the EL system to generate a set of test data by linking named entities in our NER test data to Wikipedia pages. The English Wikipedia snapshot was dumped in April 2014
which contains around 4.6M pages. Using this method we generate a training data set with 4,699 English Wikipedia pages and a test set of 415 English Wikipedia pages.

Notice that the automatically generated classifier training and test data are \emph{weakly labeled} since the EL system may link an entity to a wrong Wikipedia page and thus the entity type assigned to that page could be wrong. Since the test data is crucial for evaluating the classification accuracy, we manually corrected the output.

\subsubsection{Classifier Performance}

To evaluate the prediction power of different types of features, we train a number of classifiers using only title features, only infobox features, only text features, and all features respectively. We show the $F_1$ score of the classifiers on different entity
types in Table \ref{table:classifier}. ALL is the overall performance, and PER (PERSON), ORG (ORGANIZATION), GPE, TITL (TITLEWORK), FAC (FACILITY) are the top five most frequently entity types in the test data.

\begin{table}
\footnotesize \centering
\begin{center}
\begin{tabular}{|c|c|c|c|c|c|c|}

\hline \textbf{Features} & \textbf{ALL} & \textbf{PER} & \textbf{ORG} & \textbf{GPE} & \textbf{TITL} & \textbf{FAC} \\
\hline Title & 62.4 & 73.4 & 67.2 & 59.0 & 57.1 & 47.1 \\
\hline Infobox & 77.3 & 92.6 & 87.8 & 92.0 & 95.4 & 50.0 \\
\hline Text & 87.2 & 97.5 & 87.3 & 95.1 & 88.5 & 40.0 \\
\hline All & 90.1 & 96.1 & 92.5 & 95.1 & 96.9 & 75.0 \\
\hline

\end{tabular}
\end{center}
\caption{$F_1$ score of English Wikipedia entity type classifiers.} \label{table:classifier}
\end{table}

From Table \ref{table:classifier}, we can see that text features are the most important features for classifying Wikipedia pages, since the classifier trained with only text features achieves an overall $F_1$ score of 87.2, which is better than the classifier trained with either title or infobox features alone. Nevertheless, both infobox and title features provide additional useful information for entity type classification, and the classifier trained with all the features achieves an overall $F_1$ score of 90.1.

\subsubsection{Improvement via Self-Training}

Self-training is a \emph{semi-supervised} learning technique that can be used in applications where there is only a small number of labeled training examples but a large number of unlabeled examples. Since our weakly annotated classifier training data only covers around 1\% of all the Wikipedia pages, we are motivated to use self-training to further improve the classification accuracy.

We first apply a standard self-training approach. The classifier trained with the initial training data is used to decode (i.e., classify) all the unlabeled Wikipedia pages to predict their entity types with confidence scores. We add the self-decoded Wikipedia pages with high confidence scores to the training data and train a new classifier. Via experiments a threshold of 0.9 is used to sort
out high-confident self-decoded examples. The $F_1$ score of the new classifier is improved to 91.1, as shown in Table \ref{table:self-training}.

Under the standard approach, about 2.3M self-decoded examples are added, the size of which is about 500 times of the size of the original training data. The errors of the original classifier could be amplified with such a big increase of the training size with so
many self-decoded examples.

To address this issue, we have developed a \emph{sampling-based} self-training approach. Instead of adding all the self-decoded examples with confidence scores greater than or equal to 0.9, we do a random sampling of those high-confident examples. We use a sampling probability $p(e) = q\cdot c(e)$, where $q$ is a sampling ratio parameter and $c(e)$ is the confidence score of example $e$. Under this approach, examples with higher confidence scores are more likely to be selected, while the total number of selected examples is controlled by the sampling ratio $q$. Via experiments we found that a small sampling ratio like $q=0.01$ yields good improvement (although the improvement is not sensitive to $q$). As shown in Table \ref{table:self-training}, the classification accuracy under the sampling-based approach is further improved to 91.8 $F_1$ score (the improvement is calculated by averaging over 5 random samples with $q=0.01$).

\begin{table}
\small \begin{center}
\begin{tabular}{|c|c|c|}

\hline \textbf{Classifier} & \textbf{Train Size} & $\mathbf{F_1}$ \\
\hline Original Classifier & 4,699 &  90.1  \\
\hline Self-Training (Standard) & +2,352,836 & 91.1 \\
\hline Self-Training (Sampling) & +26,518 & 91.8 \\
\hline
\end{tabular}
\end{center}
\caption{Improving classifier accuracy via self-training.} \label{table:self-training}
\end{table}

\subsection{Wikipedia Entity Type Mapping}

We construct an English Wikipedia entity type mapping by applying the English Wikipedia entity type classifier on all the English Wikipedia pages ($\sim$4.6M). Each entry of the mapping includes an \emph{entity name} (which is extracted from the title of a Wikipedia page) and the associated \emph{entity type} with \emph{confidence score} (which is determined by the classifier). We denote the English Wikipedia entity type mapping that includes all the pages by \emph{English-Wiki-Mapping}.

To build a high-accuracy mapping, one may want to include only entities with confidence scores greater than or equal to a threshold $t$ in the mapping, and we denote such a mapping by \emph{English-Wiki-Mapping}($t$). Notice that a mapping with a higher $t$ will have more
accurate entity types for its entities, but it will include fewer entities. Therefore, there is a trade-off between \emph{accuracy} and \emph{coverage} of the mapping, which can be tuned by the confidence threshold $t$.  There are about 2.9M entities in \emph{English-Wiki-Mapping}(0.9), which covers about 63\% of all the English Wikipedia pages.

We have also found that the \emph{length} of an entity name (i.e., number of words in an entity name) also plays an important role for determining which entities should be included in the mapping for improving an NER system. Therefore, we use \emph{English-Wiki-Mapping}($t,i$) to denote the English Wikipedia entity type mapping that includes all the entities with confidence scores greater than or equal to $t$ and at least $i$ words in their names. \emph{English-Wiki-Mapping}(0.9,2) covers about 55\% of all the English Wikipedia pages, and \emph{English-Wiki-Mapping}(0.9,3) covers about 25\% of all the English Wikipedia pages.

\section{Multilingual Wikipedia Entity Type Mapping} \label{section:multilingual_wiki_mapping}

Based on the English Wikipedia entity type mapping, we want to build high-accuracy, high-coverage Wikipedia entity type mappings for other languages with minimum human annotation and language-dependent knowledge involved. We utilize the \emph{inter-language links} of Wikipedia, which are the links between one entity's pages in different languages. The inter-language links between English Wikipedia pages and Wikipedia pages of another language provide useful information for this task.

Suppose we want to build a Wikipedia entity type mapping for a new language, and we use Portuguese as an example. A direct approach is \emph{projection} using the inter-language links between English and Portuguese Wikipedia pages: for each Portuguese Wikipedia page that has an inter-language link to an English Wikipedia page, we project the entity type of the English Wikipedia page (determined by the
English entity type mapping) to the Portuguese Wikipedia page. The rationale is that both the English and Portuguese pages are describing the same entity, even probably with different spelling (e.g., \textbf{United States} in English vs. \textbf{Estados Unidos} in Portuguese), the entity type of that entity does not change from one language to another.

However, the main limitation of the direct projection approach is coverage. Only a fraction of all the Portuguese Wikipedia pages have inter-language links to English Wikipedia pages, and among those pages only a subset of them have classified entity types with confidence scores high enough (e.g., at least 0.9). For example, projecting \emph{English-Wiki-Mapping}(0.9) to Portuguese Wikipedia returns 143K pages, which covers only 15\% of all the Portuguese Wikipedia pages (around 920K in total).

We apply an alternative approach, which uses the 143K Portuguese Wikipedia pages (acquired by projection from \emph{English-Wiki-Mapping}(0.9)) as weakly annotated training data to train a Portuguese Wikipedia entity type classifier. For feature engineering purpose, we also project the English Wikipedia entity type classifier training and test data (as described in Section
\ref{section:eng-data}) to Portuguese Wikipedia pages via inter-language links, and this produces 1,190 Portuguese Wikipedia pages which are used as the test data. Pages in the test data set are excluded from the 143K training data set.

We use similar features (title, infobox and text) as for the English classifiers to train the Portuguese classifiers. Again we find that the classifier trained with all the features achieves the best accuracy of 86.3 $F_1$ score. Notice that this is an approximated evaluation
because the pages in the test data set are labeled via projection and not by human.

We build Portuguese Wikipedia entity type mappings by applying the Portuguese Wikipedia entity type classifier on all the Portuguese Wikipedia pages. We use \emph{Portuguese-Wiki-Mapping}($t$) to denote the mapping that includes entities with confidence scores greater than or equal to a threshold $t$. There are 525K entities in \emph{Portuguese-Wiki-Mapping}(0.9), which covers about 57\% of all the Portuguese Wikipedia pages, a significant improvement of coverage compared to the direct projection approach (15\%).

The main advantage of our approach is that no human annotation or language-dependent knowledge is required, so it can be easily applied to a new language. We have applied this approach to build high-accuracy, high-coverage Wikipedia entity type mappings for several new languages including Portuguese, Japanese, Spanish, Dutch and German.

\section{Improving NER Systems}

We have developed several approaches that utilize the Wikipedia entity type mappings to improve NER systems. Let $\mathcal{M}$ be a Wikipedia entity type mapping. For an entity name $x$, let $\mathcal{M}(x)$ denote the set of possible entity types for $x$ determined by the mapping. If an entity name $x$ is in the mapping, then $\mathcal{M}(x)$ includes at least one entity type, i.e.,
$|\mathcal{M}(x)| \geq 1$, where $|\mathcal{M}(x)|$ is the cardinality of $\mathcal{M}(x)$. Otherwise if an entity name $x$ is not in the mapping, then $\mathcal{M}(x)=\emptyset$ is the empty set and $|\mathcal{M}(x)| =0$.

The first approach is to use a Wikipedia entity type mapping $\mathcal{M}$ as a \emph{decoding constraint} for an NER system. Under this approach, the mapping is applied as a constraint during the decoding procedure: if a sequence of words in the text form an entity name $x$ that is included in the mapping, i.e., $|\mathcal{M}(x)| \geq 1$, then the sequence of words will be identified as an entity, and its entity type is determined by the decoding algorithm while being constrained to one of the entity types in $\mathcal{M}(x)$.

The second approach is to use a Wikipedia entity type mapping $\mathcal{M}$ to \emph{post-process}  the output of an NER system. Under this approach, the mapping is applied after the decoding procedure: if the name of a system entity $x$ is in the mapping and the entity type for that entity name is unique based on the mapping, i.e., $|\mathcal{M}(x)| = 1$, then its entity type will be determined by the unique entity type in $\mathcal{M}(x)$.

The decoding constraint approach is more aggressive than the post-processing approach, because it may create new entities and change entity boundaries. This approach is more reliable for entities with longer names. Via experiments we find that using \emph{Wiki-Mapping}(0.9,2) or \emph{Wiki-Mapping}(0.9,3) achieves the best improvement under the decoding constraint approach. Remember \emph{Wiki-Mapping}($t,i$) includes all the entities with confidence scores at least $t$ and at least $i$ words in their names.

In contrast, the post-processing approach is a more conservative approach since it relies on the system entity boundaries and only changes their entity types if determined by the mapping, so it will not create new entities. Via experiments we find that using \emph{Wiki-Mapping}(0.9,2) achieves the best improvement under the post-processing approach.

Based on the observation that the decoding constraint approach is more reliable for longer entities while the post-processing approach can better handle short entities, we have designed a \emph{joint} approach that combines the two approaches as follows: it first applies \emph{Wiki-Mapping}(0.9,3) as a decoding constraint for an NER system to produce system entities, and then applies \emph{Wiki-Mapping}(0.9,2) to post-process the system output.  The joint approach combines the advantages of both approaches and achieves robust performance in our experiments.

Finally, we can use a Wikipedia entity type mapping to create \emph{dictionary features} for training an NER system. The idea of using Wikipedia to create training features was explored before, e.g., \cite{kazama07,ratinov09,radford15}. The difference between our approach and the previous approaches is how the features are created: we first build high-accuracy, high-coverage multilingual Wikipedia entity type mappings and then use the mappings to generate dictionary features. Via experiments we find that using \emph{Wiki-Mapping}(0.9,1) or \emph{Wiki-Mapping}(0.9,2) achieves the best improvement under the dictionary feature approach.

\section{Experiments}

In this section, we evaluate the effectiveness of the proposed Wikipedia-based approaches via experiments on NER systems trained for 6 languages: English, Portuguese, Japanese, Spanish, Dutch and German. For each language, we compare the baseline NER system with the following approaches:
\begin{itemize}

\item DC($i$): the decoding constraint approach with mapping \emph{Language-Wiki-Mapping}(0.9,$i$).

\item PP($i$): the post-processing approach with mapping \emph{Language-Wiki-Mapping}(0.9,$i$).

\item Joint: the joint approach that combines DC(3) and PP(2).

\item DF($i$): the dictionary feature approach with mapping \emph{Language-Wiki-Mapping}(0.9,$i$).
\end{itemize}

To evaluate the generalization capability of an NER system, we compute the $F_1$ score on the unseen entities (\emph{Unseen}) as well as on all the entities (\emph{All}) in a test data set.

\subsection{English}

The baseline English NER system is a CRF model trained with 328K tokens of human-annotated news articles. It uses standard NER features in the literature including $n$-gram word features, word type features, prefix and suffix features, Brown cluster type features, gazetteer features, document-level cache features, etc.

We have two human-annotated test data sets: the first set, Test (News), consists of 40K tokens of human-annotated news articles; and the second set, Test (Political), consists of 77K tokens of human-annotated political party articles from Wikipedia. The results are shown in Table \ref{table:klue-eng-results}.

For Test (News) which is in the same domain as the training data, the baseline system achieves 88.2 $F_1$ score on all the entities, and a relatively low $F_1$ score of 78.7 on the unseen entities (38\% of all the entities are unseen entities). The dictionary feature approach DF(2) achieves the highest $F_1$ scores among the Wikipedia-based approaches. It improves the baseline system by 1.2 $F_1$ score on all the entities and by 3.1 $F_1$ score on the unseen entities. The joint approach achieves the second highest $F_1$ scores. It improves the baseline by 0.7 $F_1$ score on all the entities and by 2.0 $F_1$ score on the unseen entities.

For Test (Political) which is in a different domain from the training data, the fraction of unseen entities increases to 84\%. In this case, the $F_1$ score of the baseline system drops to 64.1, and the Wikipedia-based approaches demonstrate larger improvements. For example, DF(2) improves the baseline system by 2.7 $F_1$ score on all the entities and by 3.6 $F_1$ score on the unseen entities.

\begin{table}
\begin{center}

\begin{tabular}{|c|c|c|c|c|}

\hline \textbf{NER} & \multicolumn {2}{|c|}{\textbf{Test (News)}}  & \multicolumn {2}{|c|}{\textbf{Test (Political)}}  \\
\cline{2-5}   \textbf{System}   & \emph{All }& \emph{Unseen} & \emph{All} & \emph{Unseen}  \\
   & 100\% & 38\% & 100\% & 84\%  \\
\hline Baseline & 88.2 & 78.7   & 64.1  &  60.9      \\
\hline DC(2) & 88.1 & 79.4 & 66.3  & 63.5\\
\hline DC(3) & 88.7 & 80.2  & 65.8 & 62.9 \\
\hline PP(2) & 88.6  & 79.8 & 64.7  & 61.7\\
\hline Joint & 88.9 & 80.7   & 66.3 & 63.6   \\
\hline DF(1) & 88.5 & 80.0 & 66.3 & 64.2 \\
\hline DF(2) & \textbf{89.4} & \textbf{81.8} & \textbf{66.8} & \textbf{64.5} \\
\hline

\end{tabular}
\end{center}
\caption{Experimental results for English NER (the highest $F_1$ score among all approaches in a column is shown in bold).}
\label{table:klue-eng-results}
\end{table}

\subsection{Portuguese}

For Portuguese, we have applied a semi-supervised learning approach to build the baseline NER system. The training data set includes 31K tokens of human-annotated news articles, and 2M tokens of weakly annotated data. The weakly annotated data is generated as follows. We have a large number of parallel sentences between English and Portuguese news articles. We apply the English NER system on the English sentences and project the entity type tags to the Portuguese sentences via alignments between the English and Portuguese sentences.

The baseline NER system is an MEMM model (CRF cannot handle such a big size of training data, since our NER system has 51 entity types, and the number of features and training time of CRF grow at least quadratically in the number of entity types). The test data set consists of 34K tokens of human-annotated Portuguese news articles.

The results are shown in Table \ref{table:klue-pt-results}. Because the system is trained with little human-annotated training data, the performance of the baseline system achieves only 60.1 $F_1$ score on all the entities and 50.2 $F_1$ score on the unseen entities (80\% of all the entities). In this case, the more aggressive decoding constraint approach DC(2) achieves the best improvement among the
Wikipedia-based approaches, which improves the baseline by 5.9 $F_1$ score on all the entities and by 8.6 $F_1$ score on the unseen entities. The joint approach improves the baseline by 3.0 $F_1$ score on all the entities and by 4.3 $F_1$ score on the unseen entities.

\begin{table}
\begin{center}
\begin{tabular}{|c|c|c|}

\hline \textbf{NER} & \multicolumn {2}{|c|}{\textbf{Test (News)}}   \\
\cline{2-3}  \textbf{System }   & \emph{All} & \emph{Unseen}   \\
             & 100\% &  80\% \\
\hline Baseline & 60.1 & 50.2     \\
\hline DC(2) & \textbf{66.0} & \textbf{58.8} \\
\hline DC(3) & 62.2 & 53.4 \\
\hline PP(2) & 60.9 & 51.4 \\
\hline Joint &  63.1 & 54.5  \\
\hline DF(1) & 62.4 & 52.7 \\
\hline DF(2) & 61.3 & 51.9 \\
\hline

\end{tabular}
\end{center}
\caption{Experimental results for Portuguese NER.} \label{table:klue-pt-results}
\end{table}

\subsection{Japanese}

For Japanese, the baseline NER system is an MEMM model trained with 20K tokens of human-annotated news articles and 2.1M tokens of weakly annotated data. The weakly annotated data was generated using similar steps as for the Portuguese NER system. The test data set consists of 22K tokens of human-annotated Japanese news articles.

The results are shown in Table \ref{table:klue-ja-results}. Again, in this low-resource case, DC(2) achieves the best improvement among the Wikipedia-based approaches. It improves the baseline by 9.0 $F_1$ score on all the entities and by 18.3 $F_1$ score on the unseen entities (59\% of all the entities). The joint approach improves the baseline by 4.8 $F_1$ score on all the entities and by $9.6$ $F_1$ score on the unseen entities.

\begin{table}
\begin{center}
\begin{tabular}{|c|c|c|}

\hline \textbf{NER }& \multicolumn {2}{|c|}{\textbf{Test (News)}}   \\
\cline{2-3}  \textbf{System}    & \emph{All} & \emph{Unseen}  \\
   & 100\% &  59\% \\
\hline Baseline & 50.8 & 27.3  \\
\hline DC(2) & \textbf{59.8} & \textbf{45.6} \\
\hline DC(3) & 55.6 & 36.9 \\
\hline PP(2) & 50.8 & 27.3   \\
\hline Joint &  55.6 & 36.9  \\
\hline DF(1) & 52.9 & 29.0 \\
\hline DF(2) & 51.8 & 28.0 \\
\hline

\end{tabular}
\end{center}
\caption{Experimental results for Japanese NER.} \label{table:klue-ja-results}
\end{table}

\subsection{Spanish, Dutch and German}

We also evaluate the Wikipedia-based approaches on Spanish, Dutch and German NER systems trained with the CoNLL data sets \cite{sang02,sang03}. There are only 4 entity types in the CoNLL data: PER (person), ORG (organization), LOC (location), MISC (miscellaneous names). Accordingly, we have trained a CoNLL-style Wikipedia entity type classifier that produces the CoNLL entity types.
The training data for the classifier is generated by using the CoNLL English training data set and the AIDA-YAGO2 data set that provides the Wikipedia titles for the named entities in the CoNLL English data set \cite{hoffart11}. Applying the classifier on all the English Wikipedia pages, we construct a CoNLL-style English Wikipedia entity type mapping. We then build CoNLL-style Wikipedia entity type mappings for Spanish, Dutch and German using steps as described in Section \ref{section:multilingual_wiki_mapping}.

For each of the three languages, the baseline NER system is a CRF model trained with human-annotated news data ($\sim$200K tokens), and there are two test data sets, TestA and TestB, that are also human-annotated news data (ranging from 40K to 70K tokens). The results are shown in Table \ref{table:conll-deu-results}. For Dutch and German, DF(1) achieves the best improvement among the Wikipedia-based approaches. For Spanish, the joint approach achieves the best improvement among the Wikipedia-based approaches. Again, in all cases, the Wikipedia-based approaches demonstrate larger improvements (ranging from 1.0 to 3.4 $F_1$ score) on the unseen
entities.

\begin{table}
\begin{center}

\begin{tabular}{|c|c|c|c|c|}

\hline \textbf{NER }& \multicolumn {2}{|c|}{\textbf{TestA}}  & \multicolumn {2}{|c|}{\textbf{TestB}}  \\
\cline{2-5}   \textbf{System}   & \emph{All} & \emph{Unseen} & \emph{All} & \emph{Unseen}  \\
\hline \hline
 \textbf{Spanish}   & 100\% &  47\% & 100\% & 38\%  \\
\hline Baseline & 77.9 & 69.4  & 81.5  & 71.0 \\
\hline DC(2) & 77.9 & 69.7  & 81.4 & 71.0 \\
\hline DC(3) & 78.4 & 70.1  & 81.6 & 71.2\\
\hline PP(2) & 78.2  & 70.1 & \textbf{82.0} & \textbf{72.1}\\
\hline Joint & \textbf{78.5} & \textbf{70.4}   & \textbf{82.0} & \textbf{72.1}   \\
\hline DF(1) & 77.7 & 69.6 & \textbf{82.0} & 71.6 \\
\hline DF(2) & \textbf{78.5} & \textbf{70.4} & 81.4 & 70.9 \\
\hline \hline
 \textbf{Dutch} & 100\% & 60\% & 100\% & 54\%  \\
\hline Baseline & 80.7 & 70.8   & 82.3  &  70.9      \\
\hline DC(2) & 80.8 & 71.3  & 82.8 &  71.9 \\
\hline DC(3) & 80.8 & 71.2   & 82.4 & 71.1 \\
\hline PP(2) & 81.2 & 71.6 & 83.2 & 72.5  \\
\hline Joint & 81.3 & 71.9  & 83.1 & 72.3   \\
\hline DF(1) & \textbf{82.3} & \textbf{73.2} & \textbf{84.5} & \textbf{74.3} \\
\hline DF(2) & 81.1 & 71.1 & 83.3 & 72.5 \\
\hline \hline
\textbf{German}   & 100\% & 72\% & 100\% & 70\%  \\
\hline Baseline & 69.6 & 63.0  & 70.3  &  63.0     \\
\hline DC(2) & 70.1  & 63.8  & 70.1   & 62.8 \\
\hline DC(3) & 69.9 &  63.5  & 70.4 & 63.1   \\
\hline PP(2) & 70.5 & 64.4 & 70.6 &  63.4 \\
\hline Joint & 70.8 & 64.8  & 70.6 & 63.4   \\
\hline DF(1) & \textbf{71.8} & \textbf{65.8} & \textbf{71.8} & \textbf{65.3} \\
\hline DF(2) & 71.2 & 65.4 & 70.5 & 63.6 \\
\hline

\end{tabular}
\end{center}
\caption{Experimental results for Spanish, Dutch, and German NER.} \label{table:conll-deu-results}
\end{table}

\subsection{Discussion}

From the experimental results, we have the following observations:

\begin{itemize}
\item NER systems are more likely to make mistakes on unseen entities. In all cases, the $F_1$ score of an NER system on all the entities is always higher than the $F_1$ score on the unseen entities.

\item The Wikipedia-based approaches are effective in improving the generalization capability of NER systems (i.e., improving the accuracy on unseen entities), especially when a system is applied to a new domain (3.6 $F_1$ score improvement on political party articles/English NER) or it is trained with little human-annotated training data (18.3 $F_1$ score improvement on Japanese NER). 

\item In the low-resource scenario where an NER system is trained with little human-annotated data (e.g., 20K-30K tokens of training data for the Portuguese and Japanese systems), the decoding constraint approach, which uses a high-accuracy, high-coverage Wikipedia entity type mapping to create constraints during the decoding phase, achieves the best improvement.

\item In the rich-resource scenario where an NER system is well trained (e.g., 200K-300K tokens of training data for the English, Dutch and German systems), the dictionary feature approach, which uses a Wikipedia entity type mapping to create dictionary type features, achieves the best improvement.

\item In both scenarios, the joint approach, which combines the decoding constraint approach and the post-processing approach in a smart way, achieves relatively robust performance among the Wikipedia-based approaches.

\end{itemize}

\section{Conclusion}
In this paper, we proposed and evaluated several approaches that utilize high-accuracy, high-coverage Wikipedia entity type mappings to improve multilingual NER systems. These mappings are built from weakly annotated data, and can be easily extended to new languages with no human annotation or language-dependent knowledge involved.

Experimental results show that the Wikipedia-based approaches are effective in improving the generalization capability of NER systems. When a system is well trained, the dictionary feature approach achieves the best improvement over the baseline system; while when a system is trained with little human-annotated training data, a more aggressive decoding constraint approach achieves the best improvement. The improvements are larger on unseen entities, and the approaches are especially useful when a system is applied to a new domain or it is trained with little training data.

\section*{Acknowledgments}
We would like to thank Avirup Sil for helpful comments, and for collecting the Wikipedia data. We also thank the anonymous reviewers for their suggestions.

\bibliographystyle{emnlp2016}
\bibliography{wiki}

\end{document}